\documentclass[runningheads]{llncs}

\usepackage{eccv}

\usepackage{eccvabbrv}

\usepackage{graphicx}
\usepackage{booktabs}

\usepackage[accsupp]{axessibility}  

\usepackage{hyperref}

\usepackage{orcidlink}
\usepackage{amsfonts} 
\usepackage{amsmath}
\usepackage{amssymb}%
\usepackage{color}            %
\usepackage{xcolor}    %
\usepackage{multirow}
\usepackage{adjustbox}
\usepackage{wrapfig}
\usepackage{caption} %
\usepackage[misc]{ifsym}
\usepackage{colortbl}
\usepackage{tabularx}
\usepackage{pifont}
\usepackage{tikz,dcolumn,subcaption}
\usepackage{comment}
\usepackage{colortbl}
\usepackage{algorithmicx}
\usepackage{algorithm}
\usepackage{algpseudocode} 
\definecolor{turquoise}{cmyk}{0.65,0,0.1,0.3}
\definecolor{purple}{rgb}{0.65,0,0.65}
\definecolor{dark_green}{rgb}{0, 0.5, 0}
\definecolor{orange}{rgb}{0.8, 0.6, 0.2}
\definecolor{red}{rgb}{0.8, 0.2, 0.2}
\definecolor{darkred}{rgb}{0.6, 0.1, 0.05}
\definecolor{blueish}{rgb}{0.0, 0.3, .6}
\definecolor{light_gray}{rgb}{0.8, 0.8, 0.8}
\definecolor{pink}{rgb}{1, 0, 1}
\definecolor{greyblue}{rgb}{0.25, 0.25, 1}
\definecolor{mistyrose}{rgb}{1.0, 0.89, 0.88}
\definecolor{whitee}{rgb}{1.0, 1.0, 1.0}
\definecolor{palerobineggblue}{rgb}{0.59, 0.87, 0.82}
\definecolor{lavenderblue}{rgb}{0.9, 0.9, 1.0}
\definecolor{headerblue}{RGB}{220,230,242}
\definecolor{bestrow}{RGB}{232,245,233}
\definecolor{groupgray}{RGB}{245,245,245}
\usepackage{soul}          
\usepackage{ulem}          

\begin{document}

\title{Inverse Neural Operator for ODE Parameter Optimization} 

\titlerunning{Inverse Neural Operator for ODE Parameter Optimization}

\author{Zhi-Song Liu\inst{1,2}\orcidlink{0000-0003-4507-3097} \and
Wenqing Peng\inst{2,3} \and
Helmi Toropainen\inst{2,3} \and 
Ammar Kheder\inst{1,2} \and 
Andreas Rupp\inst{4} \and 
Holger Fröning\inst{5} \and 
Xiaojie Lin\inst{6} \and 
Michael Boy\inst{1,2,3}}
\authorrunning{Z.-S.~Liu et al.}
\institute{Department of Computational Engineering, Lappeenranta-Lahti University of Technology, Lahti, Finland \and
Atmospheric Modelling Center Lahti (AMC-Lahti), Finland \and
University of Helsinki, Helsinki, Finland \and
Saarland University, Saarbrücken, Germany \and
University of Heidelberg, Heidelberg, Germany \and
Zhejiang University, Hangzhou, China\\
\email{zhisong.liu@lut.fi}}

\maketitle

\begin{abstract}
We propose the Inverse Neural Operator (INO), a two-stage framework for recovering hidden
ODE parameters from sparse, partial observations. In Stage 1, a Conditional Fourier Neural
Operator (C-FNO) with cross-attention learns a differentiable surrogate that reconstructs
full ODE trajectories from arbitrary sparse inputs, suppressing high-frequency artifacts via spectral regularization. In Stage 2, an Amortized Drifting Model (ADM) learns a
kernel-weighted velocity field in parameter space, transporting random parameter
initializations toward the ground truth without backpropagating through the surrogate,
avoiding the Jacobian instabilities that afflict gradient-based inversion in stiff regimes. Experiments on a real-world stiff atmospheric chemistry benchmark (POLLU, 25 parameters) and a synthetic Gene Regulatory Network (GRN, 40 parameters) show that INO outperforms gradient-based and amortized baselines in parameter recovery accuracy while requiring only $\sim$0.23\,s inference time, a $487\times$ speedup over iterative gradient descent.
  \keywords{Neural Operator \and ODE inversion \and chemical modeling}
\end{abstract}

\section{Introduction}
\begin{figure*}[t]
	\centering
		\centerline{\includegraphics[width=1\columnwidth]{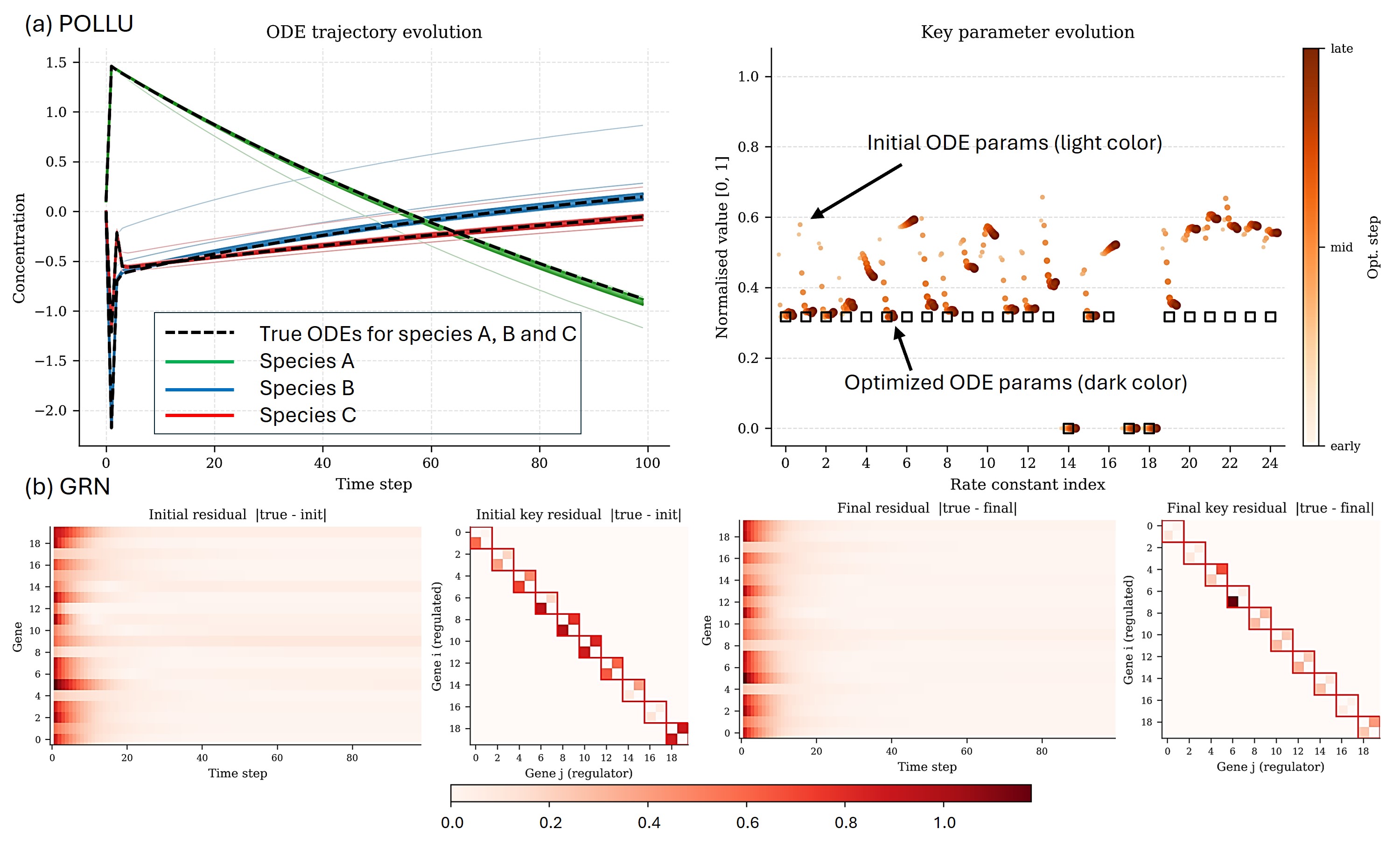}}
		\caption{\small{\textbf{ODE parameter optimization via the proposed Inverse Neural Operator (INO).}
		INO recovers hidden ODE parameters from sparse partial observations across two benchmarks.
		\textbf{(a) POLLU} (chemical kinetics): 25 unknown reaction rate coefficients governing 20 chemical species.
		Parameters evolve from random initialization (light orange) toward the ground truth (dark orange),
		while the predicted trajectory simultaneously converges to the true ODE solution.
		\textbf{(b) GRN} (gene regulatory network): 40 active regulatory coefficients within a $20\times20$
		interaction matrix. Residual heatmaps show that both the recovered parameters and the predicted
		gene expression trajectories converge to near-zero error after optimization.
		}}
		\label{fig:teaser}
\end{figure*}

The mathematical modeling of dynamical systems through Ordinary Differential Equations (ODEs) is a cornerstone of modern scientific inquiry. Gene Regulatory Networks (GRNs) encode transcriptional interactions that govern cell fate and disease~\cite{davidson,karlebach}; industrial chemical reactors are designed and controlled via kinetic ODE models whose parameters determine yield and safety~\cite{fogler}; and fluid dynamics at continuum scales is governed by PDE systems whose reduced-order surrogates are themselves ODE-like~\cite{brunton2020machine}. In all these domains, the inverse problem, inferring optimal system parameters from observed trajectories, is critical for accelerating scientific discovery, optimizing industrial production, and enabling robust uncertainty quantification~\cite{tarantola2005inverse}.

Despite its importance, parameter inference is inherently ill-posed. In real-world scenarios, researchers rarely possess the luxury of dense, noise-free, or full-state observations. Instead, they must contend with three compounding difficulties. Observational Sparsity: measurements are often restricted to a handful of discrete time points, a regime where classical adjoint methods become unreliable~\cite{rackauckas2020universal}. Partial Observability: only a subset of the system's components can be monitored, leaving latent dynamics hidden. Experimental Noise: stochastic fluctuations and measurement errors obfuscate the underlying physical manifold~\cite{calderhead2009accelerating}. Traditional approaches rely on iterative optimization-in-the-loop or adjoint-based sensitivity analysis~\cite{chen2018neural,rackauckas2020universal}. While mathematically grounded, these methods are computationally prohibitive for stiff systems~\cite{kim2021stiff}, highly sensitive to initialization~\cite{raue2013joining}, and frequently trapped in non-convex loss landscapes~\cite{transtrum2011geometry}. Furthermore, while Neural Operators such as FNO~\cite{fno1} have emerged as powerful forward surrogates, they are prone to spectral aliasing and high-frequency distortions under sparse data, producing physically inconsistent oscillations in the recovered trajectories~\cite{wang2022improved}.

In this paper, we propose the Inverse Neural Operator (INO), a novel framework that treats
parameter recovery as an amortized generative task rather than a per-instance optimization
problem. Our approach addresses the twin challenges of ill-posedness and computational cost through two primary technical innovations. First, we introduce a Conditional Fourier Neural Operator (C-FNO) integrated with Cross-Attention, which functions as a spectral regularizer to suppress the Gibbs phenomenon~\cite{gottlieb1997gibbs} and ensure temporal coherence. The affine conditioning mechanism of C-FNO is inspired by Feature-wise Linear
Modulation~\cite{film}, adapting feature-wise scale-and-shift to the operator learning setting. Second, we employ an Amortized Drifting Model (ADM) to learn a global vector field in the parameter space, allowing for rapid parameter refinement that is significantly more stable than traditional gradient descent. Unlike standard Flow
Matching~\cite{lipman2022flow}, which still relies on backpropagation through the surrogate to compute a gradient-based supervision target, ADM constructs its training signal from a kernel-weighted drifting field built entirely from forward-pass residuals, making it fully Jacobian-free at training time (see Section~\ref{sec:method}).

Empirical evaluations on stiff chemical kinetics (POLLU~\cite{pollu}) and synthetic
biological benchmarks (GRN~\cite{grn}) demonstrate that INO achieves substantial speedups
over iterative gradient-based methods while maintaining superior parameter recovery accuracy.
Concretely, the ADM requires only 20 integration steps ($\approx$\,0.23\,s per sample)
compared to 100 gradient-descent iterations ($\approx$\,112\,s), yielding a
\textbf{${\sim}487\times$} wall-clock speedup with improved accuracy. Our main
contributions are:

\begin{itemize}
\item We formulate the Inverse Neural Operator (INO), an amortized framework that
sidesteps the instabilities of iterative optimization in high dimensional, ill-posed ODE
parameter spaces.
\item We propose a Spectral Cross-Attention mechanism within a Conditional FNO
(C-FNO) that mitigates high-frequency distortions in Fourier-based operators, enforcing
physical consistency in the learned ODE trajectories.
\item We introduce the Amortized Drifting Model (ADM), which replaces
Jacobian-based gradient supervision with a kernel-weighted residual drifting field,
transforming the inverse parameter search into a stable, Jacobian-free transport problem.
We provide theoretical analysis connecting ADM to mean-field interacting particle
systems~\cite{carrillo2010particle,liu2016svgd} and derive conditions under which the
ensemble distribution contracts toward the ground-truth parameter.
\end{itemize}


\section{Related Work}
\label{sec:relwork}
We consider that our work is closely related to two fields of research: 1) neural operators for functional learning, and 2) inverse problems for ODEs/PDEs and their optimization. For 1), we focus on different neural operator architectures for ODE/PDE modeling. For 2), we are interested in solving the ill-posedness problem so we can estimate hidden parameters given limited observation data.

\subsection{Neural Operators for ODE/PDE Modeling}
DeepONet~\cite{deeponet} is one of the pioneering works that introduces functional learning: using neural networks to learn the mapping between infinite-dimensional functional spaces. Later, Fourier Neural Operator (FNO)~\cite{fno1,fno2} uses the Fast Fourier Transform (FFT) to convert time-series data to the frequency domain, so that the time-dependent integral operation can be computed by matrix multiplication. \cite{cono} improves FNO by projecting into the complex domain for Complex Neural Operator (CNO) learning. Wavelet Neural Operator (WNO)~\cite{wno} further improves FNO by using wavelet transforms. Laplacian Neural Operator (LNO)~\cite{lno} leverages Laplace Transforms to handle non-periodic signals. GINO~\cite{gino} combines graph and Fourier architectures to learn signed distance functions for 3D PDE problems. UNO~\cite{uno} builds a hierarchical U-Net shaped neural operator for multiscale feature representation. GANO~\cite{gano} generalizes generative adversarial nets to function spaces. KANO~\cite{kano} utilizes Kolmogorov--Arnold Networks embedded in the pseudo-differential operator framework. Most recently, Continuum Attention for Neural Operators~\cite{calvello2024continuum} is proposed to formulate self-attention as a map between infinite-dimensional function spaces and proving a universal theorem for neural operators.

Our C-FNO conditioning mechanism employs affine feature modulation (scale and shift) of latent representations conditioned on ODE parameters, inspired by Feature-wise Linear Modulation~\cite{film}. While FiLM was originally developed for visual question answering, its general conditioning principle transfers naturally to operator learning, where parameter embeddings serve as the conditions.

\subsection{Inverse Problems and Parameter Inference}
The study of neural operators applies to many ODE/PDE problems, including fluid dynamics~\cite{fluid}, continuum mechanics~\cite{continuum}, weather forecasting~\cite{graphcast}, atmospheric chemistry~\cite{nne}, and astrophysics~\cite{astrophysics}. Many researchers have studied neural operators for PDE/ODE solver acceleration~\cite{speedup_1}, interpretable deep learning models~\cite{benchmark}, and discovering unknown physics~\cite{discover}. We are particularly interested in using neural operators for parameter optimization.

The computational efficiency of neural operators allows us to quickly solve ODE/PDE problems for arbitrary parameters, enabling inverse problem formulations to recover unknown ODE/PDE parameters. This is both computationally challenging with existing numerical solvers and ill-posed due to noisy measurements and limited data. Markov-Chain Monte-Carlo (MCMC) sampling is often used for inverse problems, but requires extensive forward simulation. Several works~\cite{inverse_2,inverse_3} improve on this using neural emulators. \cite{po} discusses PDE parameter optimization on several classic problems. \cite{nio} proposes a neural inverse operator (NIO) combining DeepONet and FNO for forward and backward ODE/PDE simulation. \cite{inverse} improves DeepONet via an invertible branch net. \cite{ifno} proposes joint forward and inverse optimization via invertible FNO. However, these methods require multiple stages of training and typically assume access to sufficiently informative observations.

Our work differs from these approaches in two key respects. First, rather than inverting through a fixed deterministic map, we train an ADM that learns a globally regularized vector field in parameter space, providing robustness to the ill-posedness inherent in sparse-observation settings. Second, by eliminating Jacobian computation from the inversion stage, our method avoids the sensitivity collapse that afflicts gradient-based inversion through stiff neural operators.

\section{Method}
\label{sec:method}
\begin{figure*}[t]
	\centering
		\centerline{\includegraphics[width=1\columnwidth]{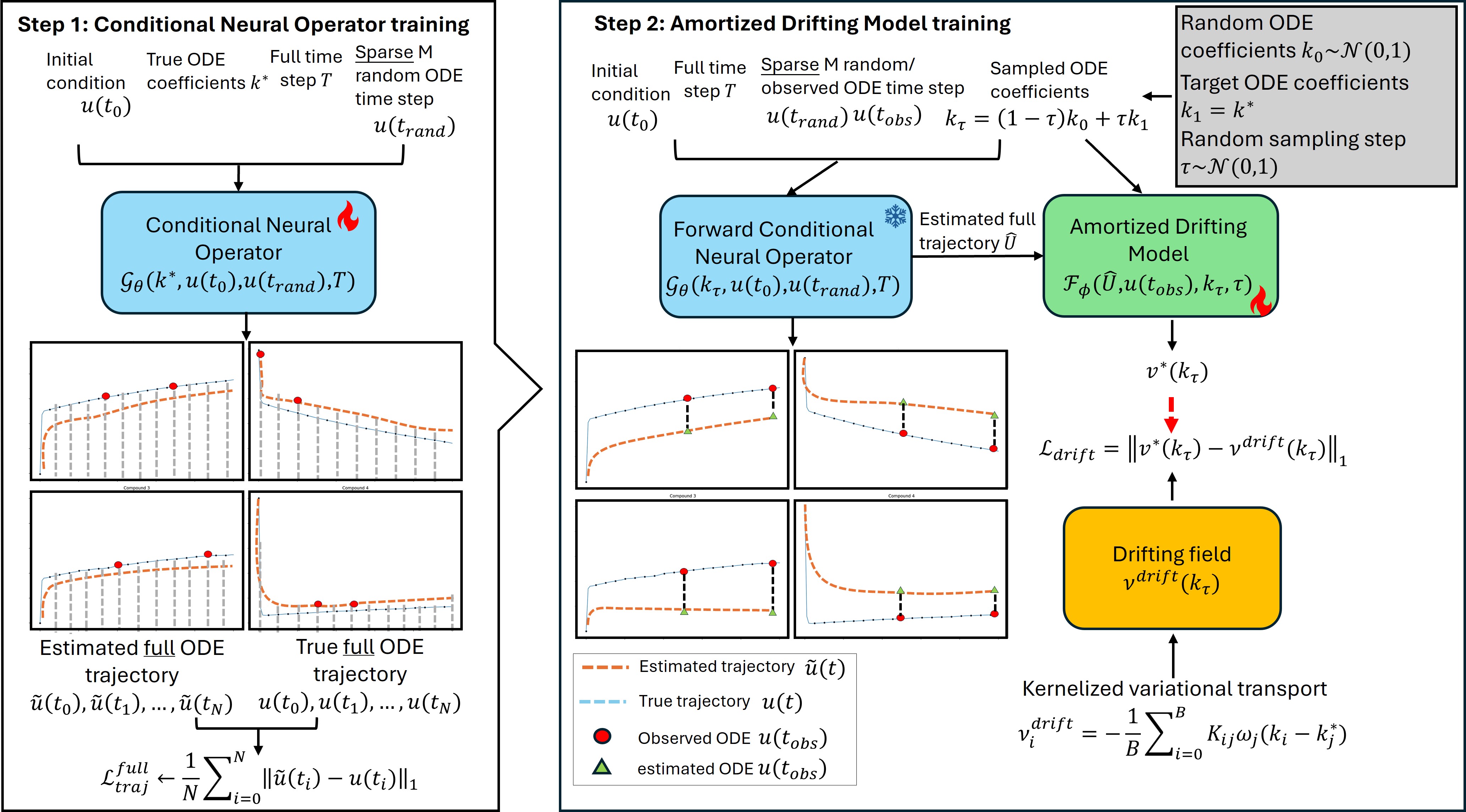}}
		\caption{\small{\textbf{Overall architecture of the proposed Inverse Neural Operator (INO).} INO decouples forward surrogate learning from inverse parameter recovery across two stages. Stage 1 (CNO): a Conditional FNO with affine parameter modulation and Cross-Attention reconstructs the full ODE trajectory from sparse partial observations. Stage 2 (ADM): the frozen CNO acts as a forward evaluator only; pairwise residuals drive a kernel-weighted drifting field that transports random parameter initializations toward the ground truth without backpropagating.
        }}
		\label{fig:network}
\end{figure*}

\subsection{Preliminary}

We begin by revisiting the definition of neural operator learning. Let $(x, y)$ be pairs of data, where $x \in \mathcal{X}$ and $y \in \mathcal{Y}$ are functions defined on a $d$-dimensional spatial domain $\Omega \subset \mathbb{R}^d$, and let $\mathcal{A}: \mathcal{X} \rightarrow \mathcal{Y}$ denote an operator such that $\mathcal{A}(x) = y$. The goal of neural operator learning is to find a parametric approximation $\hat{\mathcal{A}}$ of $\mathcal{A}$, such that for any new input $x' \in \mathcal{X}$ we have $\hat{\mathcal{A}}(x') \approx \mathcal{A}(x')$. In other words, we aim to learn a mapping in the functional space that generalizes to both seen and unseen inputs.

To implement this, we discretize the functions at a set of sensor points $c_1, \dots, c_m \in \Omega$ and parametrize the neural operator with a set of parameters $\theta \in \mathbb{R}^N$. The training of the neural operator can then be formulated as the following optimization problem:

\begin{small}
\begin{equation}
\min_{\theta \in \mathbb{R}^N} \sum_{(x,y) \in \text{data}} L\big(\hat{\mathcal{A}}(x;\theta), y\big),
\label{eq:no}
\end{equation}
\end{small}

\noindent where $L$ is a suitable loss function, such as mean squared error. Recovering the solution operator is challenging in practice, since $\mathcal{A}$ is often nonlinear and high-dimensional, and the available data may be scarce or noisy. One effective approach is the Fourier Neural Operator (FNO)~\cite{fno1}. The key idea is to exploit the Fast Fourier Transform (FFT) to convert integral operations with a kernel into multiplications in the frequency domain. For each local domain $\Omega_i \subset \Omega$:

\begin{small}
\begin{equation}
\int_{\Omega_i} k^{(i)}(x - y) \, u_i(y) \, dy 
= \mathcal{F}^{-1} \Big( \mathcal{F}(k^{(i)}) \cdot \mathcal{F}(u_i) \Big)(x) 
= \mathcal{F}^{-1} \Big( \mathcal{R} \cdot \mathcal{F}(u_i) \Big)(x), 
\quad x \in \Omega_i,
\label{eq:fno}
\end{equation}
\end{small}

\noindent where $k^{(i)}$ is the integral kernel, $u_i$ is the input function restricted to $\Omega_i$, $\mathcal{F}$ and $\mathcal{F}^{-1}$ denote the forward and inverse Fourier transforms respectively, and $\mathcal{R} = \mathcal{F}(k^{(i)})$ are the truncated Fourier coefficients. This replaces a dense integral with a frequency-domain convolution at quasi-linear complexity $\mathcal{O}(m \log m)$.

\subsection{Proposed INO for Hidden ODE Parameter Optimization}
As shown in Figure~\ref{fig:network}, we propose a two-stage framework for hidden ODE parameter optimization: 1) a Conditional Neural Operator (CNO) to learn the ODE surrogate, and 2) an Amortized Drifting Model (ADM) for hidden ODE parameter optimization. In step 1, the CNO takes sparse observations at random time steps and the corresponding ODE parameters to predict the full ODE trajectory. While neural operators are differentiable, differentiability alone does not guarantee stable or global inverse dynamics. Rather than relying on local sensitivity information, our method replaces direct differentiation with a learned global transport field in the parameter space. In the second stage, the trained CNO is frozen and used to supervise the ADM, which learns a kernel-weighted, Jacobian-free inverse mapping. Crucially, the ADM supervision target is constructed entirely from forward evaluations of $\mathcal{G}_{\theta^*}$ without computing any gradient through the surrogate. This is the key distinction from standard Flow Matching (see Section~\ref{subsec:adm}).

\noindent\textbf{Conditional Neural Operator Learning.} In step 1, given dense ODE trajectory points $u(t_i)$ sampled from the full time sequence $T$, and true ODE parameters $k^*$, we randomly sample $M$ time steps $t_\mathrm{rand}$ and their corresponding sparse observations $u(t_\mathrm{rand})$ as conditional input to the neural operator $\mathcal{G}(\cdot)$. The goal is to approximate the full ODE trajectory:

\begin{small}
\begin{equation}
\mathcal{L}_{traj}^{full}=\frac{1}{N}\sum_{i=0}^N ||\hat{u}(t_i)-u(t_i)||_1
\label{eq:loss_1}
\end{equation}
\end{small}

\noindent where $\hat{u}(t_i)$ is the model prediction at time $t_i$. Different from existing neural operator learning, the proposed CNO: 1) learns conditional trajectory generation from known ODE parameters and partial observations, similar to parametrized extrapolation; 2) randomly samples partial ODE observations $u(t_\mathrm{rand})$ at each training iteration, so the model learns to map arbitrary partial observations to the unique ODE solution.

Note that we keep initial conditions as fixed values. The partial ODE observations are much sparser ($M\ll N$), making the problem ill-posed. To improve ODE approximation, we combine the Fourier Neural Operator~\cite{fno1} with cross-attention~\cite{attn} to construct the network.

\begin{figure*}[t]
	\centering
		\centerline{\includegraphics[width=1\columnwidth]{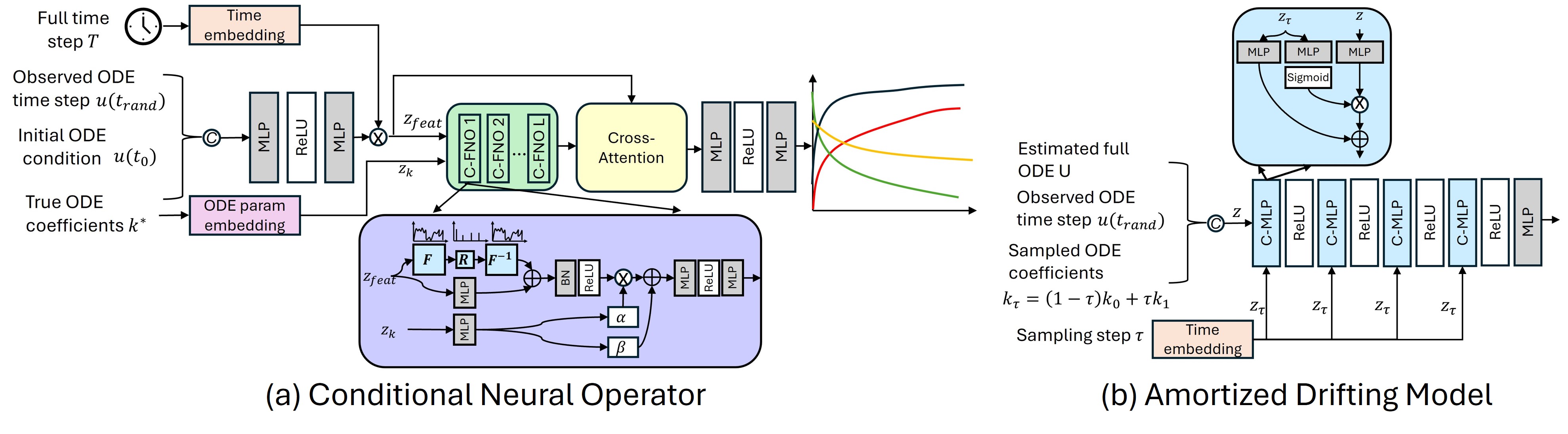}}
		\caption{\small{\textbf{Overall architecture of the proposed components.} Left: the Conditional Neural Operator (CNO), consisting of conditional FNO (C-FNO) blocks and a Cross-Attention block producing the full ODE solution given ODE parameters and partial observations. Right: the Amortized Drifting Model (ADM), consisting of conditional MLP blocks (C-MLP) that learn a kernel-weighted drifting velocity field in parameter space, supervised without backpropagation through the surrogate.
		}}
		\label{fig:architecture}
\end{figure*}

The left subfigure of Figure~\ref{fig:architecture} shows the structure of the proposed CNO. The key idea is to concatenate observed ODE time steps, initial ODE values, and ODE parameters, then project them to the latent space. A novel conditional FNO (C-FNO) block learns correlations between partial observations and ODE parameters in the latent space. Cross-attention then supervises the full trajectory before and after the C-FNO process. Finally, MLP layers project the latent vector back to the spatial domain.

Mathematically, let $z_{feat}$ denote the latent representation of the initial ODE trajectory, and $z_k$ the embedded ODE parameters. The C-FNO block learns a conditional parametrized ODE process as:

\begin{small}
\begin{equation}
\begin{aligned}
c &= BN\Big(\mathcal{F}^{-1} \Big( \mathcal{R} \cdot \mathcal{F}(z_{feat}^i) \Big)+f(z_{feat}^i)\Big) \\
z_{feat}^{i+1} &= f\big(\max(c,0)\cdot(1+\alpha(z_k))+\beta(z_k)\big)
\label{eq:cno}
\end{aligned}
\end{equation}
\end{small}

\noindent where $z_{feat}^i$ is the $i$-th input ODE latent feature and $BN$ is batch normalization. The affine terms $\alpha(z_k)$ and $\beta(z_k)$ implement feature-wise scale-and-shift conditioning on the ODE parameters, following the Feature-wise Linear Modulation (FiLM) paradigm~\cite{film}. This allows the FNO's spectral operations to be globally conditioned on the ODE parameter embedding. We stack $L$ C-FNO blocks to obtain the conditional ODE feature $z_{feat}^L$. Next, we apply cross-attention to compute the long-term time dependency between the initial ODE feature $z_{feat}^0=z_{feat}$ and the optimized feature $z_{feat}^L$:

\begin{small}
\begin{equation}
z=z_{feat}^L + \mathrm{softmax}\big(\frac{Q(z_{feat}^0)K(z_{feat}^L)}{\sqrt{d}}\big)V(z_{feat}^L)
\label{eq:ca}
\end{equation}
\end{small}

\noindent where $d$ is the latent dimension and $z$ is the final optimized ODE latent code, projected back to the spatial domain via MLP layers. The cross-attention uses $z_{feat}^0$ as query and $z_{feat}^L$ as key-value, so the initial (unmodified) feature acts as a reference that retrieves temporally coherent structure from the C-FNO-optimized representation. Intuitively, this acts as a residual spectral regularizer: the C-FNO blocks learn to approximate the time integral via truncated FFT, introducing potential high-frequency artifacts, while the cross-attention compares the evolved representation against the clean initial feature and suppresses non-physical oscillations. As seen from Figure~\ref{fig:traj}(a), cross-attention effectively enforces temporal coherence, producing smooth, physically consistent trajectories.

\noindent\textbf{From Gradient-Based Inversion to Jacobian-Free Drifting.}
\label{subsec:adm}
Given a pre-trained Conditional Neural Operator $\mathcal{G}_{\theta^*}$, a natural strategy for inverse parameter recovery is to fix $\theta^*$ and optimize the ODE parameter $k$ using partial trajectory observations:

\begin{small}
\begin{equation}
\begin{aligned}
\mathcal{L}_{traj}^{partial}&=\frac{1}{M}\sum_{i=0}^M \left\|\mathcal{G}_{\theta^*}(k^i,u(t_0),u(t_i),T)-u(t_i)\right\|_1, \\
k^{i+1}&=k^i-\lambda \nabla_{k^i}\mathcal{L}_{traj}^{partial}.
\label{eq:gd}
\end{aligned}
\end{equation}
\end{small}

\noindent By the chain rule, $\nabla_k \mathcal{L}=\left(\frac{\partial \mathcal{G}_{\theta^*}}{\partial k}\right)^{\!\top}\nabla_y \mathcal{L}$, which requires evaluating the Jacobian of the neural operator with respect to the ODE parameters.
For stiff or multi-scale dynamical systems, this Jacobian is typically highly anisotropic and ill-conditioned, resulting in unstable gradients and poor conditioning of the inverse problem.
Moreover, since the loss is computed only on sparse observations $u(t_{\text{obs}})$, the inverse mapping is inherently ill-posed: multiple parameter configurations may generate nearly identical partial trajectories.
Consequently, direct gradient descent often suffers from slow convergence, sensitivity collapse, and local minima.

One could alternatively train a standard Flow Matching model~\cite{lipman2022flow} that learns a velocity field in parameter space. However, the canonical Flow Matching formulation still requires computing the gradient $-\nabla_{k_\tau}\mathcal{L}$ through $\mathcal{G}_{\theta^*}$ to obtain the supervision target, inheriting the same Jacobian instabilities described above.

\noindent\textbf{Amortized Drifting Model (ADM).}
To eliminate Jacobian dependence entirely, we propose the ADM. Rather than supervising the flow model with a gradient-based target $v^* = -\nabla_k \mathcal{L}$, ADM replaces this signal with a kernel-based drifting field constructed purely from forward evaluations of $\mathcal{G}_{\theta^*}$, requiring no backpropagation through the surrogate at any point during training.

Let $R_i = \hat{U}_i - U_i$ denote the trajectory residual of sample $i$, where $\hat{U}_i = \mathcal{G}_{\theta^*}(k_i)$ is computed from a forward pass only. We define a similarity kernel in observation space:

\begin{equation}
K_{ij}=\exp\left(-\frac{\|R_i - R_j\|^2}{\sigma}\right),
\end{equation}

\noindent where $\sigma$ is an adaptive bandwidth (median heuristic: $\sigma = \text{med}^2 / \log B$). Samples with similar residual structures are strongly coupled. Let $w_j = \|R_j\|_2$ denote a residual magnitude weight. For a mini-batch $\{k_i\}_{i=1}^B$, the drifting velocity for ODE parameters $k_i$ is:
\begin{equation}
v_i^{\text{drift}}=-\frac{1}{B}\sum_{j=1}^BK_{ij}\,w_j\,(k_i - k_j^*).
\label{eq:drifting}
\end{equation}

This formulation requires no backpropagation through $\mathcal{G}_{\theta^*}$ and no Jacobian evaluation. Parameter updates are driven entirely by residual similarity in observation space. We finally train $\mathcal{F}_\phi$ to approximate $v^{\text{drift}}$:
\begin{equation}
\mathcal{L}_{drift}=\left\|\mathcal{F}_\phi(k_\tau, \hat{U}, u_{\text{obs}}, \tau)-v^{\text{drift}}(k_\tau)\right\|_1,
\end{equation}
and perform inference by integrating the learned vector field.

\noindent\textbf{Theoretical Analysis of ADM.}
The ensemble update in Eq.~(\ref{eq:drifting}) defines an interacting particle system. As $B \to \infty$, the empirical measure $\rho_t^B = \frac{1}{B}\sum_i \delta_{k_i(t)}$ converges to a continuous distribution $\rho_t$ satisfying the continuity equation: $\frac{\partial \rho}{\partial t} + \nabla_k \cdot \bigl(\rho\, v[\rho]\bigr) = 0$
with the nonlocal velocity field:
\begin{equation}
v[\rho](k,t) = -\int K\!\bigl(R(k), R(k')\bigr)\,\|R(k')\|\,(k - k'^{\,*})\,\rho(k',t)\,dk'.
\end{equation}
This resonates to a McKean--Vlasov-type aggregation equation~\cite{carrillo2010particle}. Standard propagation-of-chaos results~\cite{sznitman1991topics} guarantee that the finite-particle system approximates the mean-field limit in the Wasserstein distance.

The structure of $v_i^{\text{drift}}$ shares the kernel-weighted particle interaction of Stein Variational Gradient Descent (SVGD)~\cite{liu2016svgd}, but differs in two important respects: (1) the kernel $K_{ij}$ acts in residual space rather than parameter space, enabling similarity-based coupling across the observation manifold; (2) the repulsive kernel-gradient term of SVGD is absent, making ADM a consensus scheme designed for optimization rather than posterior sampling. This connects ADM more closely to Ensemble Kalman Inversion~\cite{iglesias2013ensemble}, which uses empirical cross-covariance as a global preconditioner. The drifting velocity points toward a lower residual norm at every particle location. When the kernel bandwidth $\sigma$ is large relative to inter-particle residual distances, the field approximates a global consensus pull toward $k^*$, promoting fast collective convergence. Compared to gradient descent and Flow Matching, the proposed ADM removes Jacobian dependence, introduces global mean-field regularization, and yields significantly improved stability in stiff dynamical systems.

\section{Experiments}
\label{sec:exps}
\subsection{Experimental Details}
\noindent \textbf{Datasets.}
We evaluate the proposed INO framework on two representative ODE systems:
(1) POLLU~\cite{pollu} and (2) GRN~\cite{grn}.

\noindent \textbf{POLLU.}
a widely used stiff ODE system for atmospheric chemistry modeling.
It consists of 25 chemical reactions involving 20 species, with reaction rate coefficients spanning several orders of magnitude ($10^{-3}$ to $10^{12}$), resulting in highly nonlinear and stiff dynamics.
The 25 reaction rate coefficients $\{k_i\}_{i=0}^{24}$ constitute the unknown ODE parameters to be recovered.
The detailed reaction equations are provided in the supplementary material.

\begin{figure*}[t]
	\centering
		\centerline{\includegraphics[width=1\columnwidth]{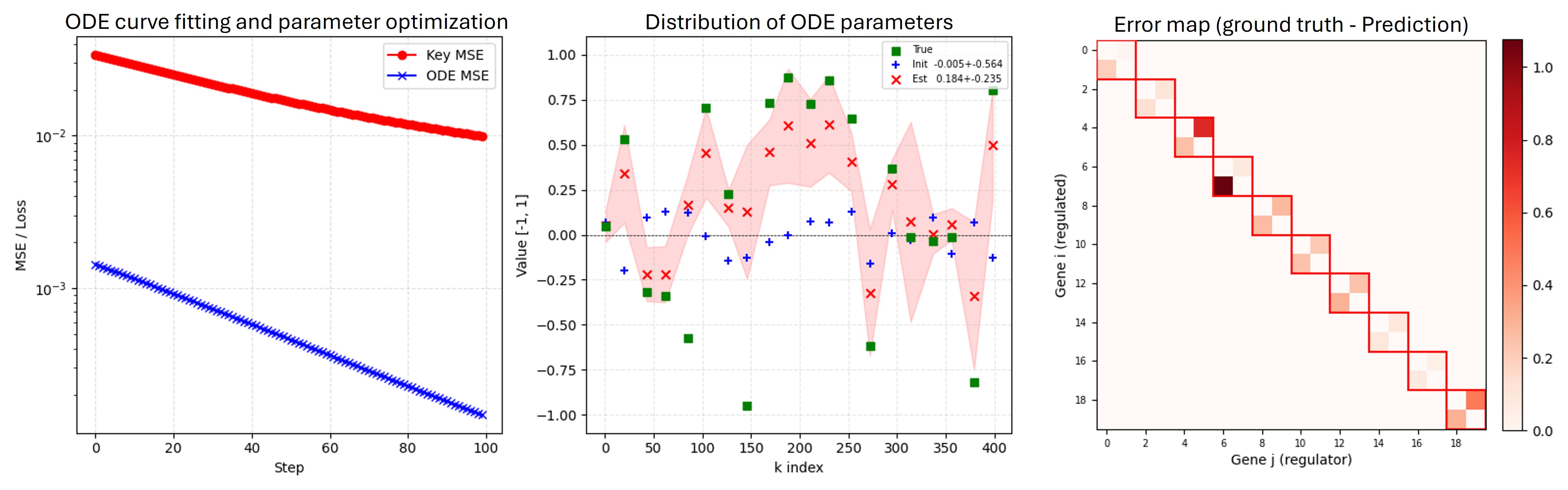}}
		\caption{\small{\textbf{Visualization of ODE parameter optimization on the GRN dataset.} Left: MSE losses on ODE fitting and parameter updating over iterations. Middle: distribution of 40 parameters after optimization. Right: residual map between ground truth and final prediction.
		}}
		\label{fig:grn}
\end{figure*}

\noindent \textbf{GRN.}
The GRN dataset models regulatory interactions among $n$ genes through a nonlinear ODE system:
$\frac{dx(t)}{dt} = c + K g(x(t)) - \Gamma x(t),$
where $K \in \mathbb{R}^{n\times n}$ denotes the interaction matrix, $c \in \mathbb{R}^{n}$ the basal transcription rate, and $\Gamma \in \mathbb{R}^{n}$ the decay rate. We fix the decay and basal rates and focus on estimating the 40 activated (diagonal and off-diagonal) entries of $K$. Specifically, in line with the GRN biological prior that regulatory networks are sparse, we activate the main diagonal and its immediate off-diagonals, yielding 40 unknown ODE parameters consistent with the GRN figure (Fig.~\ref{fig:grn}) and ablation studies. The interaction coefficients are sampled from $\mathcal{N}(0,I)$ to generate diverse regulatory regimes. Additional implementation details are provided in the supplementary material.

\noindent \textbf{Data Generation.} We fix initial conditions and vary only the ODE parameters. For each system, we generate 50,000 training samples using Latin Hypercube Sampling (LHS) to ensure uniform coverage of the parameter space (1,000 held-out test samples for all evaluations). Each trajectory is simulated for 100 time steps, producing observations $y \in \mathbb{R}^{n \times 100}$. ODE parameters are linearly normalized to $[0,1]$, while trajectory data are standardized to zero mean and unit variance. Both stages of INO are trained for 1000 epochs. The learning rate is $1\times10^{-3}$ for Stage 1 (CNO) and $1\times10^{-4}$ for Stage 2 (ADM). Batch size is 32. All experiments are conducted using PyTorch on a single NVIDIA V100 GPU.

\noindent \textbf{Evaluation.} We assess performance from both the forward modeling and inverse recovery perspectives. \textit{ODE trajectory reconstruction:} we measure the discrepancy between predicted and ground-truth trajectories using MSE and MAE. \textit{Parameter recovery:} we compute MSE and MAE between estimated and ground-truth ODE parameters. For stochastic inference procedures, we additionally report the mean and variance of the recovered parameters to quantify uncertainty.

\begin{table*}[t]
\caption{
    \textbf{Comparison with state-of-the-art methods for ODE parameter optimization.}
    We report ODE parameter recovery (Mean, Std, MAE) and trajectory fitting (MSE, MAE)
    on POLLU (25 params) and GRN (40 params).
    \textbf{Gradient}: whether backpropagation through the surrogate is required at inference.
    \colorbox{bestrow}{Green}: our method. \textbf{Bold}: best per column.
}
\centering
\setlength{\tabcolsep}{5pt}
\renewcommand\arraystretch{1.25}
\resizebox{\linewidth}{!}{
\begin{tabular}{l c cc cc cc cc cc cc}
\toprule

\multirow{3}{*}{\textbf{Method}}
& \multirow{3}{*}{\textbf{Grad.}}
& \multicolumn{5}{c}{\textbf{POLLU} (25 parameters)}
& \multicolumn{5}{c}{\textbf{GRN} (40 parameters)} \\

\cmidrule(lr){3-7} \cmidrule(lr){8-12}

& & \multicolumn{3}{c}{\textit{Param. Recovery}}
  & \multicolumn{2}{c}{\textit{Traj. Fitting}}
  & \multicolumn{3}{c}{\textit{Param. Recovery}}
  & \multicolumn{2}{c}{\textit{Traj. Fitting}} \\

\cmidrule(lr){3-5} \cmidrule(lr){6-7}
\cmidrule(lr){8-10} \cmidrule(lr){11-12}

& & Mean$\downarrow$ & Std$\downarrow$ & MAE$\downarrow$
  & MSE$\downarrow$ & MAE$\downarrow$
  & Mean$\downarrow$ & Std$\downarrow$ & MAE$\downarrow$
  & MSE$\downarrow$ & MAE$\downarrow$ \\
\midrule
\rowcolor{groupgray}
\multicolumn{12}{l}{\textit{\small Gradient-based methods}} \\
Gradient Descent  & \ding{51} & 0.0218 & \textbf{0.0901} & 0.1007 & 0.0796 & 0.0277 & 0.0092 & 0.0117 & 0.0129 & 0.0206 & 0.0551 \\
SGLD              & \ding{51} & 0.0367 & 0.2221 & 0.2921 & 0.1744 & 0.1107 & 0.0112 & 0.0156 & 0.0144 & 0.0325 & 0.0612 \\
MCMC              & \ding{51} & 0.0358 & 0.2364 & 0.2895 & 0.2029 & 0.0977 & 0.0203 & 0.0255 & 0.0310 & 0.0333 & 0.0715 \\
\midrule
\rowcolor{groupgray}
\multicolumn{12}{l}{\textit{\small Gradient-free methods}} \\
CMAES             & \ding{55} & 0.0272 & 0.1923 & 0.1389 & 0.0671 & 0.0499 & 0.0189 & 0.0235 & 0.0286 & 0.0314 & 0.0720 \\
iFNO~\cite{ifno}  & \ding{55} & 0.0643 & ---    & 0.1665 & 0.0701 & 0.0855 & 0.0233 & ---    & 0.0326 & 0.0299 & 0.0613 \\
NIO~\cite{nio}    & \ding{55} & 0.0482 & ---    & 0.1587 & 0.0720 & 0.0932 & 0.0255 & ---    & 0.0311 & 0.0328 & 0.0704 \\
SPIN-ODE~\cite{spinode} & \ding{55} & 0.0794 & --- & 0.1874 & 0.0989 & 0.0886 & 0.0364 & --- & 0.0330 & 0.0341 & 0.0722 \\
Flow Matching     & \ding{55} & 0.0300 & 0.1740 & 0.1539 & 0.0973 & 0.0855 & 0.0152 & 0.0211 & 0.0201 & 0.0324 & 0.0596 \\
\midrule
\rowcolor{bestrow}
\textbf{Drifting Model (Ours)} & \ding{55}
& \textbf{0.0117} & 0.0972 & \textbf{0.1001}
& \textbf{0.0559} & \textbf{0.0207}
& \textbf{0.0084} & \textbf{0.0115} & \textbf{0.0131}
& \textbf{0.0322} & \textbf{0.0568} \\

\bottomrule
\end{tabular}
}
\label{tab:sota}
\end{table*}

\begin{figure*}[t]
	\centering
		\centerline{\includegraphics[width=1\columnwidth]{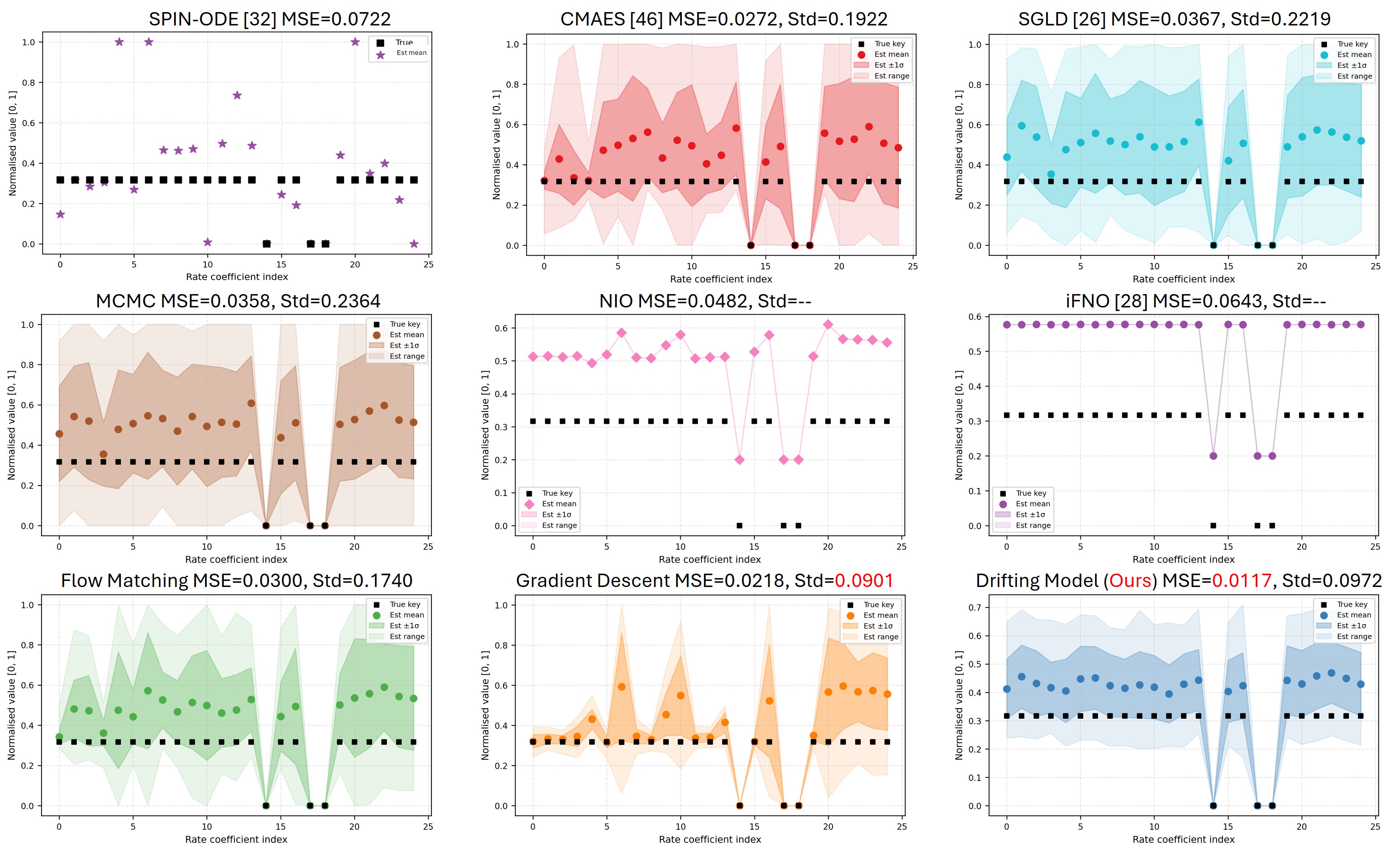}}
		\caption{\small{\textbf{Visual comparison of ODE optimization using state-of-the-art methods.} Black dots are normalized true ODE parameters. Different colors visualize the mean and variance of different methods starting from different initializations.
		}}
		\label{fig:sota}
\end{figure*}

\subsection{Comparison with State-of-the-Art}
All methods share the same pretrained CNO as their forward surrogate and are evaluated
under identical conditions: $M{=}3$ sparse observations (much fewer than the $N{=}100$
full trajectory time steps), 100 random initializations drawn from $\mathcal{N}(0,I)$,
and 100 optimization iterations where applicable.
We compare three categories of method.
\textit{Gradient-based}: Gradient Descent, Stochastic Gradient Langevin Dynamics (SGLD), and MCMC, which backpropagate through
the frozen CNO at inference.
\textit{Gradient-free optimizers}: CMAES~\cite{cmaes}, which queries the surrogate as a
black box.
\textit{Inverse operator methods}: SPIN-ODE~\cite{spinode}, iFNO~\cite{ifno}, and
NIO~\cite{nio}, which are feed-forward models trained to predict ODE parameters directly
from partial observations in a single pass.

\begin{figure*}[t]
	\centering
		\centerline{\includegraphics[width=1\columnwidth]{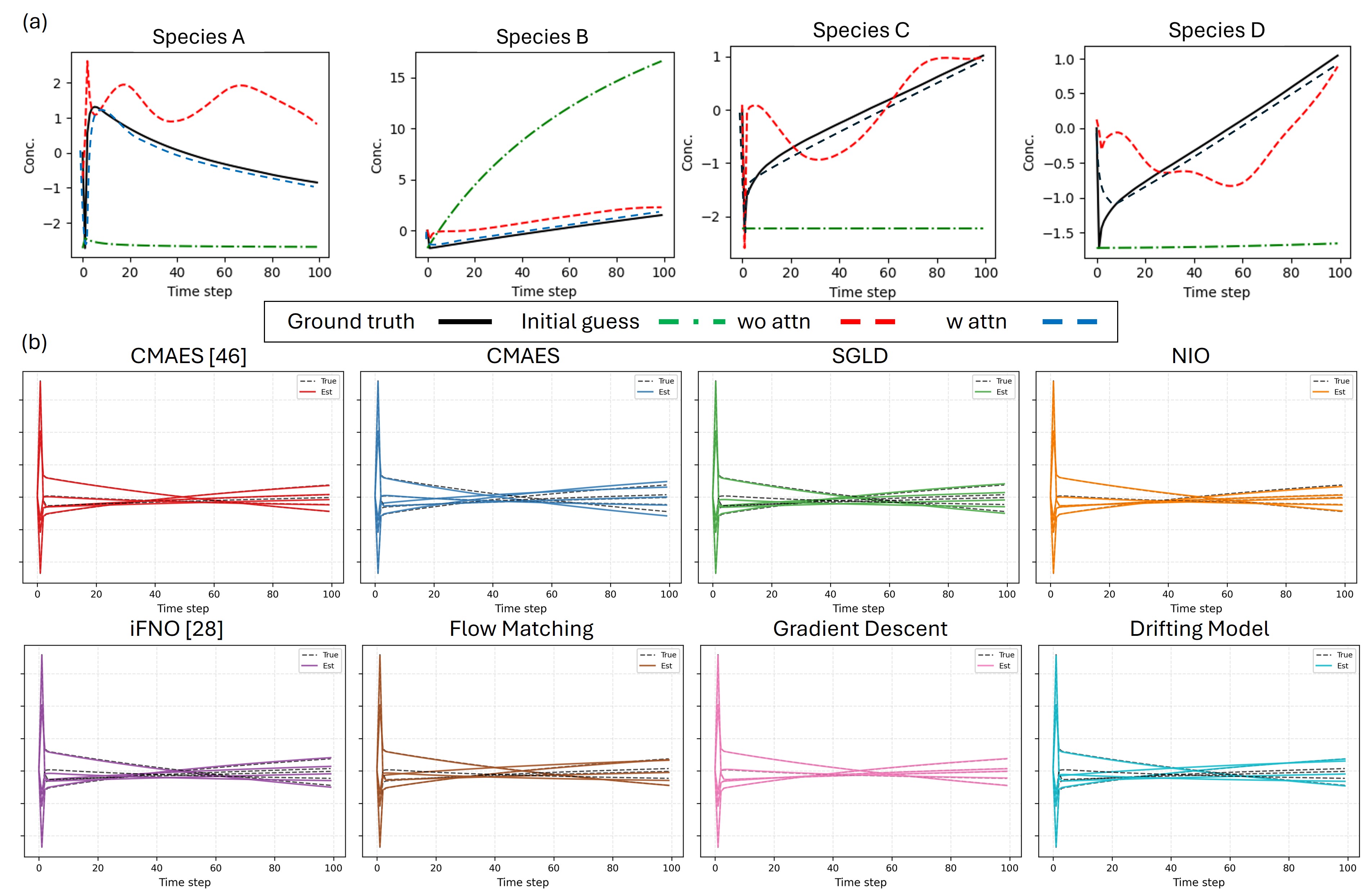}}
		\caption{\small{\textbf{Visualization of ODE trajectory fitting on the POLLU dataset.} (a) Visual comparisons of the proposed CNO with and without cross-attention. (b) ODE trajectory fitting using optimized ODE parameters from different methods (black dashed lines: true trajectories; colored lines: model estimations).
		}}
		\label{fig:traj}
\end{figure*}

Table~\ref{tab:sota} reports parameter recovery and trajectory fitting on POLLU and GRN.
Among gradient-based methods, Gradient Descent achieves the lowest standard deviation
but at the cost of 100 surrogate evaluations per sample. SGLD and MCMC introduce additional stochasticity that widens variance without improving the mean, making them poorly suited to this ill-posed setting. Among inverse operator methods, SPIN-ODE, iFNO, and NIO struggle on the parameter recovery metrics because they optimise for trajectory fit rather than parameter accuracy directly; their slightly better trajectory MSE in Figure~\ref{fig:traj}(b) is a consequence of this objective mismatch, not superior inversion quality. Our Drifting Model achieves the best mean and MAE on both datasets across all compared methods, reducing parameter mean error by $46\%$ over the next best method (Gradient Descent, $0.0218\to0.0117$ on POLLU) while requiring no backpropagation through the surrogate at inference.

Figure~\ref{fig:traj}(b) visualizes the ODE trajectories reconstructed by feeding each
method's recovered parameters back into the CNO. The Drifting Model produces trajectories that closely match the ground truth (black dashed) across all five POLLU species shown. Gradient Descent trajectories are visually competitive but exhibit occasional instability in stiff species. SPIN-ODE and NIO curves track the shape of the true ODE reasonably well, since their training directly minimizes trajectory error, yet their parameter estimates remain inaccurate, confirming that trajectory fit and parameter accuracy are not equivalent objectives in ill-posed inverse problems.

For visual comparison, Figure~\ref{fig:sota} shows statistics of every ODE parameter. We highlight $\mu\pm\sigma$ in dark color and the full range in light color. Gradient Descent shows more dramatic deviation changes, while the Amortized Drifting Model has more steady values across key parameters. Figure~\ref{fig:grn} shows the full optimization steps using ADM on the GRN dataset with 40 parameters. We see that MSE losses on ODE and parameter estimation decrease with iteration steps, and the final parameter distribution converges toward the true values.

\begin{table*}[t]
\centering
\setlength{\tabcolsep}{5pt}
\renewcommand{\arraystretch}{1.25}

\begin{minipage}{0.48\linewidth}
\centering
\caption{
    \textbf{Ablation study of CNO architecture.}
    ODE trajectory fitting (MSE$\downarrow$, MAE$\downarrow$) on POLLU and GRN.
    Each row adds one component over the previous.
    \colorbox{bestrow}{Green}: full model.
}
\resizebox{\linewidth}{!}{
\begin{tabular}{l cc cc}
\toprule
\multirow{2}{*}{\textbf{Configuration}}
& \multicolumn{2}{c}{\textbf{POLLU}}
& \multicolumn{2}{c}{\textbf{GRN}} \\
\cmidrule(lr){2-3} \cmidrule(lr){4-5}
& MSE$\downarrow$ & MAE$\downarrow$
& MSE$\downarrow$ & MAE$\downarrow$ \\
\midrule
\rowcolor{groupgray}
\multicolumn{5}{l}{\textit{\small Baseline}} \\
FNO                         & 0.1561 & 0.0894 & 0.0597 & 0.0742 \\
\midrule
\rowcolor{groupgray}
\multicolumn{5}{l}{\textit{\small Ablations (fixed observations)}} \\
C-FNO w/o Attn              & 0.0799 & 0.0689 & 0.0428 & 0.0662 \\
C-FNO w/ Attn               & 0.0715 & 0.0556 & 0.0398 & 0.0610 \\
\midrule
\rowcolor{groupgray}
\multicolumn{5}{l}{\textit{\small Ablations (random observations)}} \\
C-FNO w/o Attn + Rand       & 0.0694 & 0.0561 & 0.0401 & 0.0607 \\
\midrule
\rowcolor{bestrow}
\textbf{C-FNO w/ Attn + Rand (Ours)}
& \textbf{0.0559} & \textbf{0.0207}
& \textbf{0.0322} & \textbf{0.0568} \\

\bottomrule
\end{tabular}}
\label{tab:cno_train}
\end{minipage}
\hfill
\begin{minipage}{0.48\linewidth}
\centering
\caption{
    \textbf{Comparison of inverse optimization strategies.}
    Parameter recovery (Mean$\downarrow$, Std$\downarrow$) on POLLU and GRN. Numbers in parentheses denote optimization steps. \colorbox{bestrow}{Green}: our method.
}
\resizebox{\linewidth}{!}{
\begin{tabular}{l cc cc r}
\toprule
\multirow{2}{*}{\textbf{Method}}
& \multicolumn{2}{c}{\textbf{POLLU}}
& \multicolumn{2}{c}{\textbf{GRN}}
& \multirow{2}{*}{\textbf{Time (s)}} \\
\cmidrule(lr){2-3} \cmidrule(lr){4-5}
& Mean$\downarrow$ & Std$\downarrow$
& Mean$\downarrow$ & Std$\downarrow$ & \\
\midrule
\rowcolor{groupgray}
\multicolumn{6}{l}{\textit{\small Regression}} \\
MLP                         & 0.1023 & 0.2215 & 0.0502 & 0.0486 & 0.05 \\
\midrule
\rowcolor{groupgray}
\multicolumn{6}{l}{\textit{\small Iterative optimization}} \\
GD -- SGD (100)             & 0.0218 & \textbf{0.0901} & 0.0156 & 0.0332 & 112 \\
GD -- Adam (100)            & 0.0756 & 0.1203 & 0.0092 & 0.0117 & 114 \\
\midrule
\rowcolor{groupgray}
\multicolumn{6}{l}{\textit{\small Amortized (20 steps)}} \\
FM-Grad                & 0.0300 & 0.1741 & 0.0152 & 0.0211 & \textbf{0.21} \\
\midrule
\rowcolor{bestrow}
\textbf{Drifting Model (Ours)}
& \textbf{0.0117} & 0.0972
& \textbf{0.0084} & \textbf{0.0115} & 0.23 \\
\bottomrule
\end{tabular}}
\label{tab:cno_inverse}
\end{minipage}
\end{table*}

\subsection{Ablation Studies}
\noindent \textbf{CNO architecture design.}
Table~\ref{tab:cno_train} ablates each component of the CNO incrementally. Replacing plain FNO with C-FNO (row 1$\rightarrow$2) already yields large gains on both datasets, confirming that parameter-conditioned affine modulation~\cite{film} is essential for surrogate accuracy. Adding cross-attention (row 2$\rightarrow$3) further reduces MSE and MAE by suppressing the high-frequency artifacts introduced by truncated FFT, producing smoother and more physically consistent trajectories, see Figure~\ref{fig:traj}(a) for a visual comparison. Switching from fixed to randomly sampled observation indices at training time (row 2$\rightarrow$4) gives a comparable gain, as the operator learns to reconstruct the full trajectory from any sparse subset rather than a fixed one. Combining both (row 5, full model) achieves the best results on all four metrics, with MSE dropping from 0.1561 (FNO baseline) to 0.0559 on POLLU, a $2.8\times$ improvement.

\noindent \textbf{Effect of the Amortized Drifting Model.}
Table~\ref{tab:cno_inverse} isolates the source of ADM's gains by comparing three
paradigms: direct regression (MLP), iterative optimization (GD), and amortized
inference (FM-Grad, ADM).

Direct gradient descent achieves competitive accuracy but requires 100 iterations
($\sim$112\,s) and is sensitive to optimizer choice, SGD and Adam differ by
$3.5\times$ in mean error on POLLU, necessitating costly hyperparameter tuning.
MLP regression is the fastest (0.05\,s) but the least accurate, confirming that a
direct feed-forward mapping is insufficient for this ill-posed problem.

To isolate the contribution of the kernel-based drifting field specifically, we introduce
FM-Grad: the same $\mathcal{F}_\phi$ architecture as ADM, but trained with a
standard gradient-based target $v^* = -\nabla_{k_\tau}\|\mathcal{G}_{\theta^*}(k_\tau) -
y_\mathrm{obs}\|_1$, requiring backpropagation through the surrogate.
FM-Grad improves over plain gradient descent (0.0300 vs.\ 0.0218 mean on POLLU) while
matching its inference speed (0.21\,s), confirming that amortization alone is beneficial.
However, ADM surpasses FM-Grad by a further $2.6\times$ (0.0117 vs.\ 0.0300),
demonstrating that the performance gain is attributable specifically to the Jacobian-free
drifting supervision rather than to the shared architecture. ADM thus achieves the best accuracy across both datasets at 0.23\,s inference time, a $487\times$ speedup over iterative gradient descent with strictly higher accuracy.

\noindent\textbf{Limitations and future work.}
The current evaluation covers two ODE benchmarks; real-world measurements introduce additional challenges such as irregular sampling,
heteroscedastic noise, and partial species observability that remain to be addressed.
Future work includes validating INO on more complex real-world experimental data~\cite{calderhead2009accelerating} and extending the surrogate to PDE settings via spatiotemporal neural operators~\cite{fno1,gino}.

\section*{Conclusions}
\label{sec:conclusions}
We presented the Inverse Neural Operator (INO), a two-stage framework for recovering hidden ODE parameters from sparse partial observations. The first stage trains a Conditional Fourier Neural Operator (C-FNO) with cross-attention as a differentiable, spectrally regularized forward surrogate. The second stage trains an Amortized Drifting Model (ADM) that learns a globally regularized, Jacobian-free velocity field in the parameter manifold, supervised entirely by forward-pass residuals without any backpropagation through the surrogate. Our ablation experiments confirm that both the cross-attention mechanism and the kernel-based drifting field are essential: removing either component degrades performance, and our controlled FM-Grad comparison demonstrates that the ADM gains stem specifically from the Jacobian-free drifting supervision rather than from amortization alone. Theoretical analysis connects ADM to mean-field interacting particle systems and explains its practical robustness to observation noise via ensemble noise averaging and implicit noise marginalization during amortized training. On the POLLU and GRN benchmarks, INO achieves the best ODE parameter recovery accuracy among all compared methods while requiring only $\sim$0.23\,s inference per sample, yielding a $\sim$487$\times$ wall-clock speedup over gradient-descent baselines.

%
%
\bibliographystyle{splncs04}
\bibliography{main}

\end{document}